\documentclass[sigconf,nonacm]{acmart}  

\usepackage{tcolorbox}     
\tcbuselibrary{skins}      
\usepackage{xcolor}  
\usepackage{longtable}
\usepackage{graphicx}     
\usepackage{rotating}     

\tcbset{
  crobox/.style={
    enhanced,
    boxrule=0.5pt,
    colframe=black!40,
    colback=gray!05,
    arc=3pt,
    left=3pt,right=3pt,top=3pt,bottom=3pt,
    boxsep=0pt
  }
}

\newcommand{\dyn}[1]{\colorbox{hl}{\textbf{#1}}}

\definecolor{companiesRed}{HTML}{FF6B6B}     
\definecolor{productsBlue}{HTML}{45B7D1}      
\definecolor{inputProdGreen}{HTML}{96CEB4}    
\definecolor{inputsInputYellow}{HTML}{CCAA00} 
\definecolor{locationPurple}{HTML}{DDA0DD}    
\definecolor{industryGrey}{HTML}{BCBCBC}   
\definecolor{hl}{HTML}{FFF176}

\title[Network-KG Duality for Supply Chain Risk]{Exploring Network-Knowledge Graph Duality: A Case Study in Agentic Supply Chain Risk Analysis}
\author{Evan Heus}
\affiliation{\institution{University of California, Berkeley} \city{Berkeley, CA}\country{USA} }
\authornote{This research was conducted while Evan Heus was an MFE intern at MSCI.}
\affiliation{\institution{MSCI}\city{New York, NY}\country{USA}}
\email{evheus@berkeley.edu}

\author{Rick Bookstaber}
\affiliation{\institution{MSCI}\city{New York, NY}\country{USA}}
\email{rick.bookstaber@msci.com}

\author{Dhruv Sharma}
\affiliation{\institution{MSCI}\city{New York, NY}\country{USA}}
\email{dhruv.sharma@msci.com}

\setcopyright{none}

\usepackage{subcaption}   
\usepackage{booktabs}     
\usepackage{amsmath,amsfonts}

\begin{document}
\begin{abstract}

Large Language Models (LLMs) struggle with the complex, multi-modal, and network-native data underlying financial risk. Standard Retrieval-Augmented Generation (RAG) oversimplifies relationships, while specialist models are costly and static. We address this gap with an LLM-centric agent framework for supply chain risk analysis. Our core contribution is to exploit the inherent duality between networks and knowledge graphs (KG). We treat the supply chain network as a KG, allowing us to use structural network science principles for retrieval. A graph traverser, guided by network centrality scores, efficiently extracts the most economically salient risk paths. An agentic architecture orchestrates this graph retrieval alongside data from numerical factor tables and news streams. Crucially, it employs novel ``context shells'' --- descriptive templates that embed raw figures in natural language --- to make quantitative data fully intelligible to the LLM. This lightweight approach enables the model to generate concise, explainable, and context-rich risk narratives in real-time without costly fine-tuning or a dedicated graph database.

\end{abstract}

\maketitle

\section{Introduction}            

Large language models (LLMs) are increasingly embedded in
critical, high-stakes decision processes, from medical diagnostic triage \cite{dedhia2025bupsi} to financial advice\cite{takayanagi2025generative}. However, most domain deployments still rely on \emph{fine‑tuning} a specialist model on a
hand‑curated corpus or on straightforward
retrieval augmented generation (RAG) that proxies relationships with vector distance.  Both paths leave value on the table for supply chain risk
analysis: the former is costly to update with current events and freezes knowledge at
training time, while the latter ignores the rich \emph{network
semantics} hidden in structured data.  In addition, they
typically operate on a single modality, despite the fact that domain‑specific tasks require seamless reasoning over
text, tables, graphs, and time‑series data.

We address this gap with an LLM‑centric system that uses \emph{network science principles to uncover and expose semantically meaningful supply‑chain paths}, for example, \textbf{\textcolor{companiesRed}{Apple} $\!\leftrightarrow$ \textcolor{productsBlue}{Smartphones} $\!\leftrightarrow$ \textcolor{inputProdGreen}{Integrated Circuits}}. These distilled sub-graphs are fed into the LLM’s context at inference time. The paths are first retrieved by traversing the supply chain network from seed nodes that are semantically similar to key entities in a user's input. Using structural centrality metrics to determine traversal distance, economically relevant sub‑networks are retrieved and then formatted to expose semantic significance to the LLM. Beyond graph paths, our framework ingests
numerical factor tables and curated news snippets, fusing these
modalities into a single prompt so the model can weigh quantitative
signals alongside narrative context. This turns the graph into an ``explainable retrieval engine'' rather than a static data source. By importing only the highest‑salience paths, we give the LLM a concise, interpretable scaffold on which to reason, while freeing it
to generate natural‑language risk narratives.

\noindent\textbf{Why a network lens?}
Vector similarity treats every fact as an isolated point; supply‑chain
risk lives in the \emph{links}.  By casting the KG as a graph
\(G=(V,E)\) whose edges denote economic relations
(\textsc{Produces}, \textsc{Has Input}, \textsc{Manufactured In}),
we use network analysis to avoid over-flooding the LLM's context window.  Therefore, each retrieved path is (i) \emph{interpretable}—a concise Company → Product → Location
narrative the agent can inspect—and (ii) \emph{actionable}: the same
traversal provides quantitative signals such as edge weights for the revenue generated by a particular product for a given company.  The result is a prompt that carries
built‑in economic meaning, enabling the LLM to explain hidden
dependencies rather than guess relationships from token proximity. 

Our approach is built on a foundational insight: the duality of the supply chain network and the knowledge graph. A network is a set of nodes and edges, whereas a knowledge graph is a set of entities and semantic relationships. For supply chains, these are one and the same. The economic edge (Company A)-[PRODUCES]->(Product B) is both a structural link in a network and a semantic triple in a knowledge graph. This duality allows us to reframe the complex problem of KG traversal into an efficient network-science problem, using well-established centrality metrics to identify salient paths for LLM reasoning.

\noindent\textbf{Contributions.} We pose four main advances:
\begin{enumerate}
  \item \textbf{Network‑science path discovery} that extracts relevant sub-graphs.
  \item \textbf{Knowledge-graph semantic encoding} transforms a
        network into inference‑time prompts exposing built‑in economic meaning.
  \item \textbf{Agent‑orchestrated, multi‑modal retrieval loop}
        where a triage agent selects between graph traversal, factor
        data, and news tools prior to synthesis.
  \item \textbf{Context shells for numerical data} that wrap each
        figure in descriptive language, letting the LLM reason over context-rich
        quantitative risk metrics.
\end{enumerate}

\noindent\textbf{Paper organization.}  
Section \ref{sec:background} reviews graph-aware LLM curricula, GraphRAG, and graph database traversal.  
Section \ref{sec:system_overview} sketches the overall agentic architecture and loop.  
Section \ref{sec:data} describes the three data channels: risk factors, curated news, and synthetic supply chain graph. We also describe the tools that retrieve them, while Section \ref{sec:context_shell} zooms in on the factor \emph{context-shell} template.  
Section \ref{sec:path} explains how retrieved evidence is merged into the prompt and the rank-then-traverse algorithm that extracts salient supply chain paths.  
A live dialogue in Section \ref{sec:sample_chat} shows the retrieval chain in action, Section \ref{sec:conclusion} concludes, and the Appendix \ref{appendix} provides full relationship tables and the complete KG diagram.

\section{Background and Related Work}\label{sec:background}   
\paragraph{Bottom-Up KG curricula.}
Recent work on \emph{bottom‑up domain‑specific super‑intelligence} (BDSI) shows how
multi‑hop KG paths can be verbalized into 24,000 reasoning tasks that
supervise a 32 B‑parameter model, yielding state‑of‑the‑art scores on
ICD‑Bench \cite{dedhia2025bupsi}{}.  Their curriculum transforms the graph structure
\emph{into training data}, pushing the model to compose primitives into
higher‑order concepts.

\paragraph{Graph‑aware retrieval.}
GraphRAG \cite{edge2024graphrag} first extracts an entity knowledge
graph from a document corpus, then offline generates hierarchical
community summaries using a Leiden‑based clustering; at query time, it
assembles partial answers from those summaries and fuses them into a
global response.
This excels on corpus‑wide ``sense‑making'' questions but its
pre‑processing and multi‑stage summarization pipeline are
compute‑intensive.  Lighter variants such as Neural‑KB, GNN‑RAG\cite{mavromatis2024gnn} and
temporal‑aware RAG\cite{zhu2025fintmmbench} likewise keep the base LM frozen while injecting
graph structure to boost recall and freshness. For real-time applications such as finance, retraining such a model daily is even less feasible.


\paragraph{KG traversal in existing engines.}
Platforms such as Neo4j,\cite{neo4j_cypher_manual} and LangChain’s
Graph Retriever,\cite{bratanic2024langchain} execute a pattern like
\begin{quote}
    \centering
    \texttt{MATCH} \texttt{(a)-[:SUPPLIES*1..3]->(b)}
\end{quote}
by expanding a frontier of node
IDs from a seed vertex, applying the edge labels, hop range, property
filters, and optional \textit{top-$k$} limit that the user specifies.
Because the traversal runs inside a live Neo4j server, whose page cache
keeps hot neighborhoods in memory while the full adjacency list stays
on disk, results arrive in milliseconds. Nevertheless, they do require a running
graph service with enough RAM to cache frequently accessed portions of
the graph.  Our method retains the same query vocabulary, yet
materializes only the path segments needed at inference, avoiding the
operational overhead of a dedicated graph database while still
supporting fast, parameter-controlled look-ups.

\paragraph{Positioning of this paper.}
Our proposed agentic framework bridges the gap: it agrees with the insight in \cite{dedhia2025bupsi} that KG paths
encode domain reasoning, but applies that insight \emph{at
inference time} with lightweight network traversal, avoiding the cost
and staleness of specialist fine‑tuning while remaining faster than
on‑the‑fly graph construction methods like GraphRAG.

\section{System Overview}\label{sec:system_overview}         

Upon initiating a chat, the user supplies our system with the portfolio constituents and their weights. We make sure that the LLM holds this at the top of the context window to always be able to refer to it, since LLMs are known to get ``lost in the middle'' and not pay attention to data in the middle of the context window\cite{liu2023lostmiddlelanguagemodels}.

\begin{figure}[htbp]
  \centering
  \includegraphics[width=\linewidth]{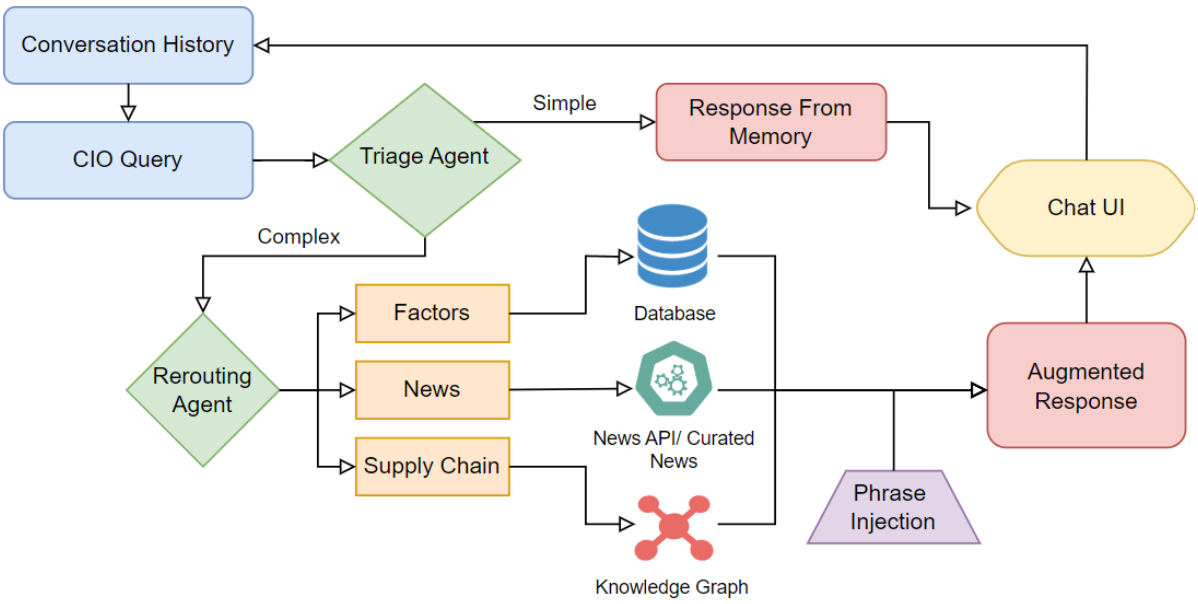}
  \caption{End‑to‑end pipeline.  The \emph{Triage Agent} decides whether
  the query can be answered from memory.  If not, a \emph{Rerouting
  Agent} selects one or more tools (Factors, News, Supply‑Chain KG),
  whose outputs are stored in a temporary database and injected into the
  LLM prompt.}
  \label{fig:arch}
\end{figure}

Figure~\ref{fig:arch} gives a bird’s‑eye view of our framework.  A user
(typically a CIO) poses a query; two lightweight agents decide 
\emph{what} to retrieve and \emph{how} to format before a `frozen' LLM
generates the final answer. Here and in what follows, by `frozen' we mean a pre-trained LLM that is not being fine-tuned as new data arrive.  

At the start of every turn, the user's message is appended to the ongoing
dialogue and sent, together with a snapshot of the portfolio, to a
lightweight \emph{Triage Agent}.  When that agent finds the answer in
memory, it returns it straight away; otherwise it flags the query for
augmentation and delegates to a downstream \emph{Rerouting Agent} that
decides which retrieval tool (factor exposures, curated news or a
supply-chain walk) to invoke and with what parameters.

The selected tools tap separate FAISS indices\cite{douze2024faiss}, one per modality, so
semantic relationships inside each data type remain intact and compound
queries\footnote{A request is termed a \emph{compound query} when it references
two or more distinct sub-topics or entities.  Embedding the full sentence
typically positions its vector between the relevant clusters in space, so
a $k$-nearest search can miss evidence dispersed across those clusters.
For example, in the query ``factor exposures and cobalt'' there may be no
single source covering both topics together, while each topic is well
covered individually at a significant vector distance from the other;
$k$-nearest retrieval therefore fails to return comprehensive coverage.} can be resolved cleanly across stores.
Each tool call is executed through the OpenAI function-calling interface; the language
model emits a JSON stub, the back-end runs the call, and the retrieved
snippets flow into a transient database before being re-injected into
the prompt.  Finally, a frozen GPT-4o model
synthesizes those multi-modal fragments with the original question,
streams an \emph{Augmented Response} to the UI, and stores the turn
in memory, readying the loop for whatever the user asks next. Responses are fast, as the tool calling system is a series of quick API calls. 

\smallskip
This modular design balances \emph{interpretability} (the network paths are
passed verbatim), \emph{flexibility} (new tools can be dropped in
without re‑training), and \emph{efficiency} (no fine‑tuning loop).  The
next section details the construction of each data mode.

\section{Data Modes And Tools}\label{sec:data}

Our system operates on three external data channels that complement the
conversation context and the user's portfolio.  Each is stored in its
own FAISS index and is retrieved through a dedicated tool.

\paragraph{Multi‑Asset‑Class (MAC) factors.}
For every security in the user's portfolio, we ingest MSCI 
MAC\cite{shepard2020msci} factor scores and the accompanying methodology text.
Before embedding, each record is wrapped in a
\emph{context shell} (see \autoref{sec:context_shell}): a short paragraph that ``tokenizes'' the numerical
z‑scores within natural‑language sentences so the LLM can reason about
their meaning rather than seeing them as opaque numbers.
The resulting embeddings power the \texttt{get\_factors} tool.

\paragraph{Curated news.}
Two news streams feed the system.  
\textbf{Macro articles} on long‑horizon risks (demographics, climate,
geopolitics) and 
\textbf{stock‑specific news} comes from the \emph{LexisNexis} archive,
restricted to the single trading day. 
All articles are chunked to roughly a page, embedded once, and served by
the \texttt{get\_news} tool; the vector store preserves outlet and
timestamp metadata to support recency filters.

\paragraph{Supply‑chain knowledge graph.}
The final mode is a multi‑entity knowledge graph that links
\textbf{\textcolor{companiesRed}{Company}}, \textbf{\textcolor{productsBlue}{Product}}, \textbf{\textcolor{inputProdGreen}{Input Product}}, \textbf{\textcolor{inputsInputYellow}{Input to Input Product}},
\textbf{\textcolor{industryGrey}{Industry}}, and \textbf{\textcolor{locationPurple}{Location}} nodes via relations such as
\textsc{Produces}, \textsc{Has Input}, \textsc{Manufactured In} and so forth (the full list of edge types are given in  \ref{tab:entity_rships_kg} in \ref{appendix}.
Although illustrative rather than fully industrial,
the graph is constructed with a pipeline akin to AIPNet \cite{fetzer2024aipnet},
to query ChatGPT's intrinsic knowledge base to create a dataset of the desired form.  These paths enter the prompt as
restructured text within a context shell, exposing the semantic relationships between nodes on the traversed graph.\\

Taken together, MAC factors, curated news, and the supply‑chain graph
give our system the multi‑modal context required for real‑time, portfolio
risk analysis.


\section{Context Shell}\label{sec:context_shell}
Domain risk analysis is based on \emph{numbers}: z-scored factor
exposures, portfolio weights, valuation ratios.  Large language
models, however, treat raw figures as opaque tokens and rarely attach contextual
economic meaning to them.  To retain a frozen, generic LLM while still
letting it reason quantitatively, our proposed system \textbf{ wraps every table row
from the MSCI Multi-Asset-Class (MAC) factor model in a
\emph{context shell}}: a short paragraph that ``tokenizes'' each figure
inside an explanatory sentence so that neighboring words endow the
number with semantics.  The surrounding text becomes part of the
embedding; the number itself is now a first-class token that via attention relationships with pertinent context becomes easily interpretable. A related but different idea has been explored in \cite{anthropic2024contextual}. 

\smallskip
\noindent\textbf{Illustrative shell.}  
Below is an excerpt from a
shell; the highlighted values are place holders for numerical figures inserted for every security whose factor exposures we have:

\begin{tcolorbox}[crobox ,width=\columnwidth]

The position in the portfolio is associated with the security
\dyn{[`Security Name']} represented by the ticker \dyn{[`Ticker']}.
This position constitutes \dyn{[`Weight']}\% of the
total portfolio.  Each of the following factors is given a \emph{z}-score
(mean 0, sd 1) for this equity relative to all other equities.\\

\smallskip\noindent
\dyn{[`Security Name']} \textbf{Equity Beta}: \dyn{[`Equity Beta']}\\
\textbf{Description:} captures market risk beyond the baseline Market factor.\\
\textbf{When High:} portfolio tilts toward high-beta stocks, amplifying risk.\\
\textbf{When Low:} portfolio tilts toward low-beta stocks, partially offsetting risk.\\

\smallskip\noindent
\dyn{[`Security Name']} \textbf{Book-to-Price}: \dyn{[`Book-to-Price']}\\
\textbf{Description:} book value divided by market capitalization.\\
\textbf{When High:} stock may be undervalued or distressed.\\
\textbf{When Low:} stock may be overvalued or considered a growth stock.
\end{tcolorbox}

When the shell is embedded and stored in a dedicated FAISS index,
both the verbal context and the numeric tokens contribute to semantic weighting.  At inference time, a \texttt{get\_factors} tool fetches the most similar shells for the
query; the LLM then attends to \emph{what the numbers mean}
rather than merely copying them, enabling factor-aware narratives
without any task-specific fine-tuning or database querying, giving the LLM a lightweight yet principled bridge between
structured factor data and natural language reasoning.

\section{Network‑Science Path Discovery}\label{sec:path}   


Here, we operationalize the network-KG duality. Instead of complex queries over a knowledge graph database, we use network traversal algorithms to retrieve economically significant sub-graphs. 

When a user query arrives, the \textbf{Triage Agent} inspects both the
conversation history and the current portfolio, deciding whether the
answer is already in memory or whether fresh data are needed.  For
queries that could benefit from the supply chain context, it delegates to a
\emph{graph-traverser} tool chosen by the downstream Rerouting
Agent.

The traverser first extracts every mention of the company, product, or location in the user’s text, embeds those strings with the same model
used to pre-compute node embeddings for the knowledge graph, and
retrieves the closest vectors; the matched vertices become
\emph{seed nodes} for traversal.
Because each vertex is stored inside a context-shell that exposes
its business metadata during embedding, similarity search is guided by both entity names and surrounding economic meaning, yielding semantically faithful starting points.

Traversal depth automatically adapts to the structural role of each
seed node. We use three network centrality measures to capture different facets of \emph{structural} node importance. \textbf{Degree} is a local score that counts immediate connections of a node,
highlighting popular ``hubs''. \textbf{Closeness} gauges how near a node lies, on average,
to \emph{all} others, rewarding globally well-connected vertices, while
\textbf{betweenness} tallies how often a node sits on shortest paths,
identifying nodes that link otherwise distant regions of the graph. These unweighted scores are utilized in this prototype as an efficient first-pass filter to identify structurally important nodes (e.g., major hubs or bottlenecks for information flow), which often proxy for economic relevance in large networks. The explicit economic weights are only applied in the subsequent semantic encoding step. 

\begin{figure}[htbp]
  \centering
  \includegraphics[width=\linewidth]{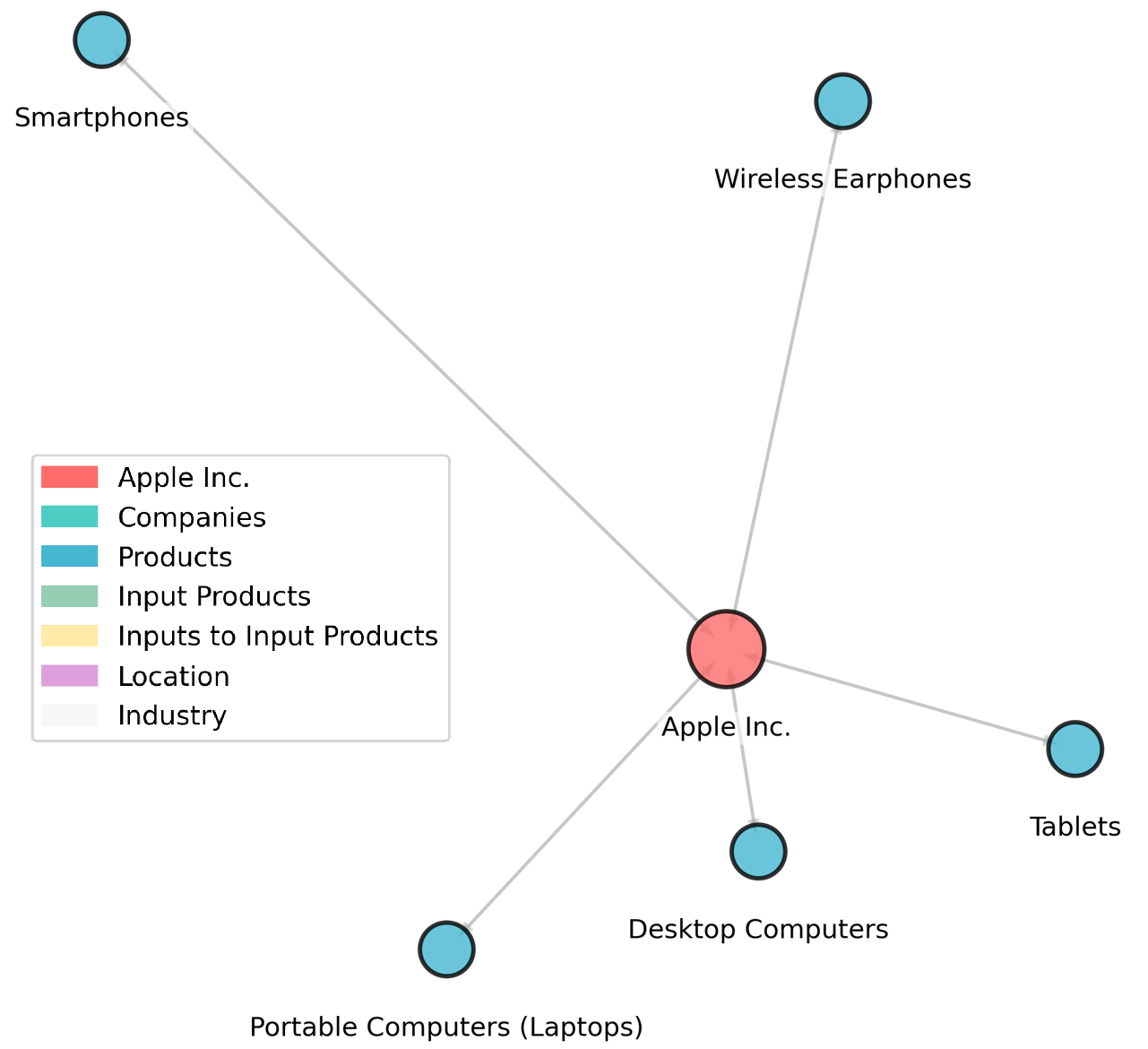}
  \caption{\textbf{\textcolor{companiesRed}{Apple's}} network with immediately adjacent \textbf{\textcolor{productsBlue}{Product}} nodes.}
  \label{fig:apple_1}
\end{figure}

\begin{figure*}[p]            
  \centering
  \includegraphics[angle=90,width=.67\textheight]{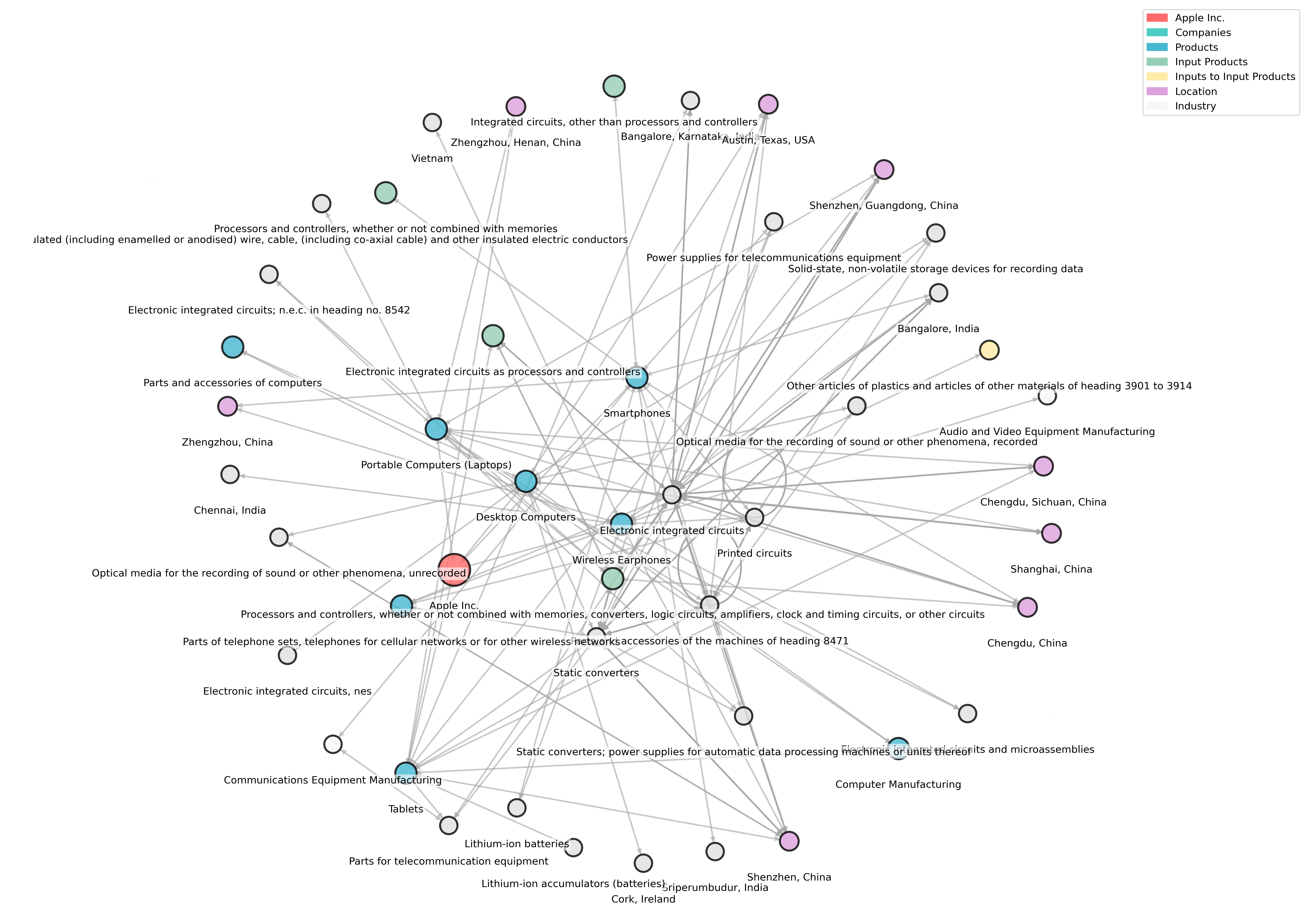}
  \caption{Two-hop supply-chain network for Apple.}
  \label{fig:apple_2}
\end{figure*}

Averaging the three yields a single salience value that reflects both the influence of a node on its neighborhood and its role in the global network
structure, giving the traversal algorithm a balanced way to decide how
far to expand from each seed node. For all nodes, this centrality metric is pre-computed. As seen when contrasting the one hop network in Figure \ref{fig:apple_1} and the two hop network in Figure \ref{fig:apple_2}, the number of nodes retrieved increases exponentially with each additional hop.

In practice supply chain networks follow a scale-free, power-law degree
distribution \cite{mungo2023reconstruct}: hubs such as \textbf{\textcolor{inputProdGreen}{Electronic Integrated Circuits}} in the middle of Figure \ref{fig:apple_2} in
the Apple network expose a wealth of information in just one hop,
whereas peripheral nodes require additional traversal hops for useful coverage.

The subtree gathered from this traverse step is then distilled
into short narrative paths. A network, strictly defined, is a collection of connected objects. Each object has minimal metadata, such as an identifier, and edges may have weights. Underneath the hood, in its most basic form, this is what a supply chain is. 

For example, if traversal identifies weighted path as 
\textbf{\textcolor{companiesRed}{Apple Inc.} $\!\rightarrow$ \textcolor{productsBlue}{Desktop Computers} $\!\rightarrow$ \textcolor{inputProdGreen}{Integrated Circuits} $\!\rightarrow$ \textcolor{locationPurple}{Shanghai, China}} with the corresponding edge weights as $10\%, 19\%, 13\%$ respectively,  our system verbalizes this structured data into an intuitive narrative.
 By treating the network as a knowledge graph, we leverage its inherent semantic structure, bringing it closer to natural language. Add node types and edge types that can be encoded into context shells, and transform this information into the following: 
\begin{tcolorbox}[crobox ,width=\columnwidth]
    \textbf{\textcolor{companiesRed}{Apple}} generates \dyn{\textbf{10\%}} of its revenue from selling
    \textbf{\textcolor{productsBlue}{Desktop Computers}}, which spends \dyn{\textbf{19\%}} of its production
    budget on \textbf{\textcolor{inputProdGreen}{Integrated Circuits}}, \dyn{\textbf{13\%}} of which are produced in
    \textbf{\textcolor{locationPurple}{Shanghai, China}}.
\end{tcolorbox}

The semantic text surrounding the injected entities and data is predefined by the edge type between two types of nodes. The complete set of edge types and metadata is provided in Appendix \ref{appendix}.

Each edge type corresponds to a semantically meaningful edge phrase, so the language model receives a compact but economically meaningful scaffold on which to reason. These supply chain snippets let AI-CRO explain \emph{why} a node matters, linking downstream cash flows to upstream inputs and vice versa depending on the CIO's query.



\section{Sample Chat}\label{sec:sample_chat}
The dialogue below illustrates how AI-CRO chains triage, tool calls, and
multi-modal retrieval in real time.  
The CIO manages a portfolio of the top-50 constituents of the MSCI USA Index, where
Apple is the largest weight, and Tesla a significant name in electric vehicles (EVs).

\begin{tcolorbox}[width=\columnwidth,colback=gray!10,colframe=gray!50,
                  title=\textbf{User}:,fonttitle=\bfseries]
\textit{I read about problems in the DRC with coltan.}
\end{tcolorbox}

\begin{tcolorbox}[width=\columnwidth,colback=yellow!10,colframe=orange!60,
                  title=\textbf{System}:,fonttitle=\bfseries]
Issues in the Democratic Republic of Congo (DRC) involving \emph{coltan},
a key source of tantalum for electronics, overlap with risks of cobalt supply that affect battery producers.  Apple and Tesla rely on both minerals, so disruptions in eastern DRC represent potential operational and
reputational vulnerabilities for the portfolio.\\[0.2em]
\textbf{Reference:} Supply-chain paths for \emph{coltan}.
\end{tcolorbox}

Our system mapped ``coltan'' to a product node, located the upstream path
into Apple’s and Tesla’s battery lines, and surfaced a portfolio-level
risk narrative.

\begin{tcolorbox}[width=\columnwidth,colback=gray!10,colframe=gray!50,
                  title=\textbf{User}:,fonttitle=\bfseries]
\textit{What has been in the news on this?}
\end{tcolorbox}

\begin{tcolorbox}[width=\columnwidth,colback=gray!25,colframe=black!65,
                  title=\textbf{Console (log)}:,fonttitle=\bfseries]
\textbf{Tool called:} \texttt{get\_news}\\
\textbf{Query embedding:} “recent news on coltan \& cobalt supply-chain
issues in the DRC’’\\
\textbf{Retrieved:} \emph{News article}, pages 3, 4 and 1
\end{tcolorbox}

\begin{tcolorbox}[width=\columnwidth,colback=yellow!10,colframe=orange!60,
                  title=\textbf{System}:,fonttitle=\bfseries]
Recent coverage documents coltan mined by the M23 militia in
eastern DRC, smuggled via Rwanda into global supply chains.  This makes
it harder for Apple, Tesla, and peers to guarantee conflict-free
sourcing, heightening both supply-disruption and ESG-reputation risk.\\[0.2em]
\textbf{Reference:} \emph{News Article}.
\end{tcolorbox}

The Rerouting Agent embedded a semantic description, retrieved the
top-three matching articles, and the LLM wove their content into the
conversation.

\begin{tcolorbox}[width=\columnwidth,colback=gray!10,colframe=gray!50,
                  title=\textbf{User}:,fonttitle=\bfseries]
\textit{Can you walk me through the various ways this could hurt Apple?}
\end{tcolorbox}

\begin{tcolorbox}[width=\columnwidth,colback=yellow!10,colframe=orange!60,
                  title=\textbf{System}:,fonttitle=\bfseries]
\textbf{(i) Supply-chain delays} — shortages of tantalum capacitors
could postpone iPhone and Mac launches.\\
\textbf{(ii) Reputational risk} — perceived reliance on conflict
minerals may erode brand trust.\\
\textbf{(iii) Margin pressure} — higher input costs compress hardware
margins and may force price rises that dampen demand.\\[0.2em]
\textbf{Reference:} Apple supply-chain paths.
\end{tcolorbox}

Across three turns, our system triaged queries, selected tools, retrieved
factor-relevant evidence, and produced narrative answers that integrate
operational, ethical, and financial dimensions, without any bespoke
fine-tuning.


\section{Conclusion and Next Steps} \label{sec:conclusion}            
\noindent\textbf{Summary.}  
Our proposed framework demonstrates that a frozen general purpose LLM can operate as a domain expert when paired with three carefully engineered data modes. MSCI MAC factor models, curated news, and a synthetic supply chain knowledge graph, along with an agent loop that decides when and how to pull each modality.  By turning high-salience graph paths into natural-language snippets and wrapping numeric factors in context shells, the system surfaces hidden dependencies and quantitative signals that ordinary vector-only RAG pipelines overlook, all while running in real time without specialist fine-tuning.  

Although our system demonstrates a powerful new approach, we acknowledge several limitations that point to avenues for future work. First, the supply chain graph was constructed using an LLM pipeline and is inherently limited in scope. While effective for this case study, it may contain inaccuracies or omissions. Future work should focus on validating and enriching this graph with structured trade data from sources such as customs declarations or bill of landing records. 

Second, our path discovery relies on unweighted topological centrality scores to guide the traversal. While this effectively identifies structurally important paths, it can overlook economically critical paths involving peripheral nodes with a high financial weight. Future work will investigate integrating edge weights directly into the traversal algorithm (e.g., using weighted shortest path algorithms or value-weighted centrality measures) to improve economic fidelity.

Finally, the system currently uses static edge weights. A key next step is to integrate real-time financial data to dynamically update these weights, reflecting changing revenue dependencies or production costs.



\bibliographystyle{ACM-Reference-Format}
\bibliography{references}

\clearpage
\onecolumn
\appendix
\section{Knowledge-Graph Details}
\label{appendix}


\begin{longtable}{| p{2.4cm} | p{6.5cm} | p{5.5cm} |}
    \caption{Relationships by Entity Type for the knowledge graph.}
    \label{tab:entity_rships_kg}\\

    \hline
    \textbf{Entity Type} & \textbf{Possible Relationships} & \textbf{Entity Metadata} \\
    \hline
    \endfirsthead
    \hline
    \textbf{Entity Type} & \textbf{Possible Relationships} & \textbf{Entity Metadata} \\
    \hline
    \endhead

    \hline
    \endfoot

    \hline
    \endlastfoot

    \textbf{Company} & 
    \textcolor{red}{\textbf{---}\textit{Produces}$\rightarrow$} Product \newline 
    \textcolor{red}{\textbf{---}\textit{Produces}$\rightarrow$} Input Product & 
    Ticker \newline Total Revenue \\ \hline

    \textbf{Product} & 
    \textcolor{red}{\textbf{---}\textit{Sold By}$\rightarrow$} Company \newline 
    \textcolor{red}{\textbf{---}\textit{Belongs To}$\rightarrow$} Industry \newline  
    \textcolor{red}{\textbf{---}\textit{Has Input}$\rightarrow$} Product (\textit{Upstream}) \newline \textcolor{red}{\textbf{---}\textit{Input To}$\rightarrow$} Product (\textit{Downstream}) \newline  
    \textcolor{red}{\textbf{---}\textit{Manufactured In}$\rightarrow$} Location \newline  \textcolor{red}{\textbf{---}\textit{Sourced From}$\rightarrow$} Location \newline
    \textcolor{red}{\textbf{---}\textit{Made With}$\rightarrow$} Input Product & 
    HS Code \newline Total Revenue Share \newline Production Cost Percentage \\ \hline

    \textbf{Industry} & 
    \textcolor{red}{\textbf{---}\textit{Includes Product}$\rightarrow$} Product \newline 
    \textcolor{red}{\textbf{---}\textit{Includes Product}$\rightarrow$} Input Product & 
    NAICS Code \\ \hline

    \textbf{Location} & 
    \textcolor{red}{\textbf{---}\textit{Production Location For}$\rightarrow$} Product & 
    Longitude \newline Latitude \newline Production Share\\
    
\end{longtable}

\begin{figure*}[h]
  \centering
  \caption{Knowledge-graph relationship diagram.}
  \includegraphics[width=\textwidth]{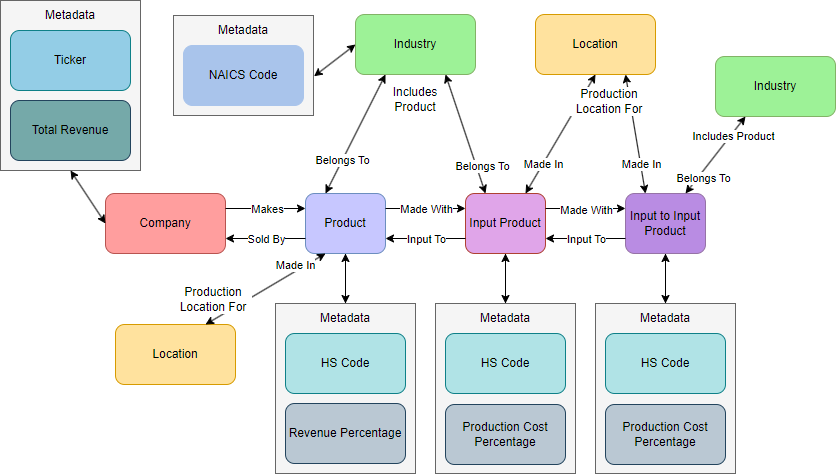}
  
  \label{fig:kg_diagram}
\end{figure*}

\end{document}